\documentclass[10pt,twocolumn]{article}

\usepackage[T1]{fontenc}
\usepackage{newtxtext}
\usepackage{newtxmath}
\usepackage[margin=0.75in,columnsep=0.25in]{geometry}
\usepackage{microtype}
\usepackage[hyphens]{url}
\usepackage{graphicx}
\usepackage{booktabs}
\usepackage{multirow}
\usepackage[table]{xcolor}
\usepackage{pifont}
\usepackage{amsmath}
\usepackage{array}
\usepackage{natbib}
\usepackage{caption}
\usepackage{float}
\usepackage{authblk}
\usepackage[hidelinks]{hyperref}

\setkeys{Gin}{keepaspectratio}
\let\cite\citep

\setcitestyle{aysep={}}
\title{\textbf{EviSD: Evidence-Conditioned Self-Distillation for Search-Augmented Agents}}

\author[1*]{Jianan Xie}
\author[2*]{Xin Sun}
\author[3]{Zhongqi Chen}
\author[3]{Xing Zheng}
\author[2]{Shu Wu}
\author[3]{Bowen Song}
\author[2]{Liang Wang}
\affil[1]{ShanghaiTech University}
\affil[2]{NLPR, MAIS, CASIA}
\affil[3]{Ant Group}
\affil[*]{Equal contribution.}
\affil[ ]{\texttt{xiejn2025@shanghaitech.edu.cn}, \texttt{xin.sun@cripic.ia.ac.cn},
          \texttt{\{shu.wu, wangliang\}@nlpr.ia.ac.cn},
          \texttt{\{chenzhongqi.czq, feishang.zx, bowen.sbw\}@antgroup.com}}
\date{}

\hypersetup{
  pdftitle={EviSD: Evidence-Conditioned Self-Distillation for Search-Augmented Agents},
  pdfauthor={Jianan Xie, Xin Sun, Zhongqi Chen, Xing Zheng, Shu Wu, Bowen Song, and Liang Wang}
}

\begin{document}

% The complete arXiv manuscript, including the supplementary material, is
% maintained directly in this file.

\maketitle

\begin{abstract}
Outcome-based reinforcement learning enables search-augmented language agents to learn from verifiable final answers, but its trajectory-level credit cannot distinguish the contributions of individual actions in a multi-turn search process. We propose \textbf{EviSD}, an evidence-conditioned self-distillation framework that uses instance-level supporting evidence as privileged information for search actions and golden answers as complementary privilege for answer actions. During training, the student samples actions from the original context, while the same model re-scores them as a privileged teacher under an action-aligned context. EviSD converts the detached teacher--student gap into a bounded correction to the outcome-derived GRPO advantage and applies it only to generated action spans. This design localizes privileged guidance while preserving the update direction determined by the outcome reward, without an auxiliary distillation objective or any change at inference time. Across seven question-answering benchmarks and three backbones spanning model scales and generations, EviSD achieves the highest macro-average Exact Match in all evaluated settings, outperforming the strongest compared methods by 1.3--2.3 points while modulating only 6.7\%--15.1\% of response tokens. \textit{Code is available at \url{https://github.com/JiananXie/EviSD}.}
\end{abstract}

% ============================================================
%   INTRODUCTION
% ============================================================
\section{Introduction}
Retrieval-augmented generation (RAG) connects language models to external evidence and has developed into a broad family of methods and evaluations~\cite{sun-etal-2025-divide,sun2026predictretrieval,du2026multimodaladaptive,chen2025competingpoisoning}. Search-augmented reasoning agents extend this paradigm by turning retrieval into a sequential decision process: they interleave internal reasoning with calls to an external retriever and decide when to search, what to query, and when enough evidence has been collected to answer~\cite{jin2025searchr1trainingllmsreason,song2025r1searcher}. Recent reinforcement learning (RL) methods make such agents trainable with verifiable final-answer rewards: the policy explores search trajectories on-policy, and GRPO converts terminal correctness into a group-relative advantage~\cite{shao2024deepseekmathpushinglimitsmathematical}. Although scalable, this feedback reveals only whether a rollout succeeded. It does not identify which searches exposed useful evidence, which generated tokens contributed to the answer, or whether a correct answer followed from an effective search strategy rather than a fortunate trajectory.

This limitation is especially acute in multi-turn search. An early targeted query can retrieve the decisive passage for all subsequent reasoning, whereas a vague or redundant query can leave the remainder of the trajectory under-informed. Nevertheless, outcome-level GRPO broadcasts the same trajectory advantage to every optimized response token. Useful queries, redundant searches, intermediate reasoning, and terminal-answer tokens therefore inherit the same outcome-derived credit, while retrieved passages are masked as environment observations. Final-answer feedback alone cannot localize the instance-specific decisions that made a trajectory effective or ineffective.

\begin{figure*}[t]
\centering
\includegraphics[width=\linewidth]{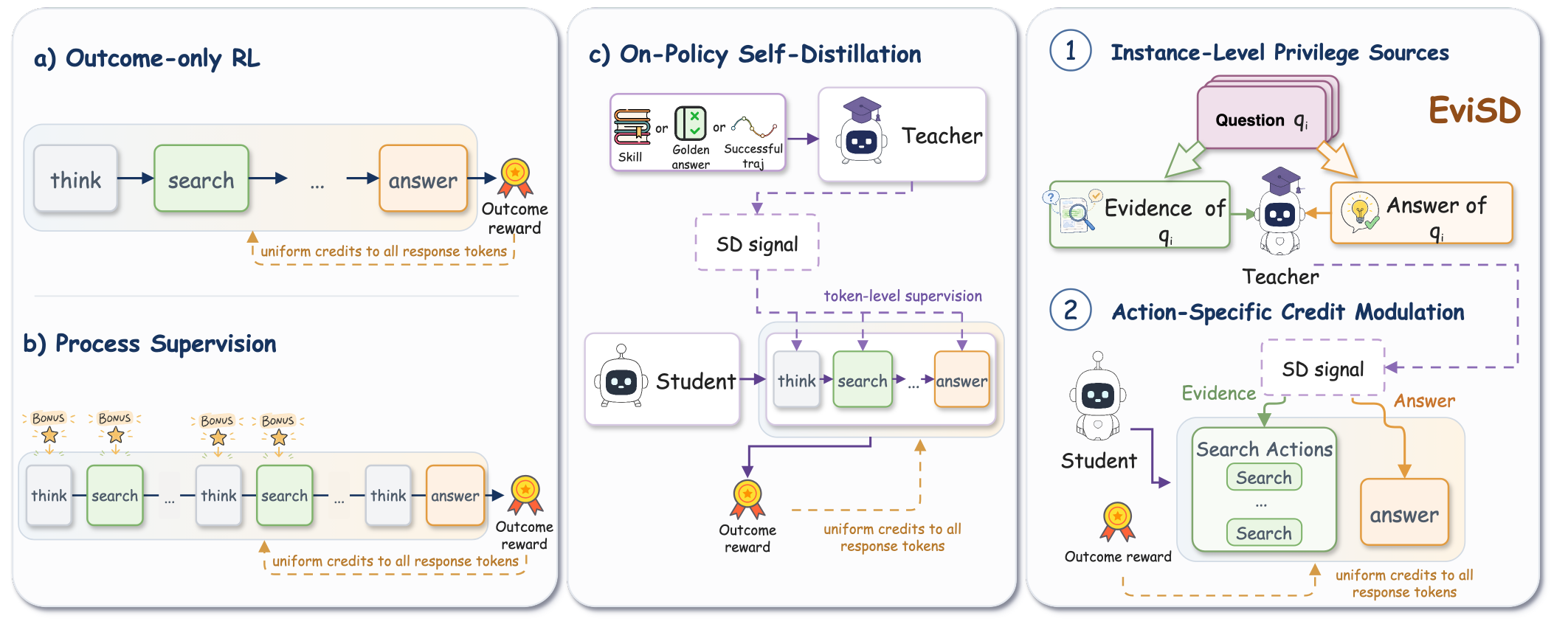}
\caption{Supervision in multi-turn search agents. Unlike outcome-only RL, process supervision, and response-wide OPSD, EviSD uses supporting evidence and golden answers as action-aligned privilege to modulate outcome credit only on search and answer action tokens.}
\label{fig:intro_teaser}
\end{figure*}

Existing methods improve credit granularity through retrieval-specific rewards or turn- and step-level estimators~\cite{shi2025searchrefinethinkfacilitating,wang2026informationgainbasedpolicyoptimization,xie2026tipsturnlevelinformationpotentialreward,feng2025groupingrouppolicyoptimizationllm,he2026hierarchyofgroupspolicyoptimizationlonghorizon}. These approaches refine when feedback is assigned, but feedback granularity alone does not determine what instance-specific information should guide a sampled search action. On-policy self-distillation (OPSD) provides a complementary mechanism: the same model samples under its inference-time context and re-scores that rollout under a privileged teacher context~\cite{zhao2026selfdistilledreasoneronpolicyselfdistillation,song2026surveyonpolicydistillationlarge}. Recent search-agent methods construct this context from task guidance, answer information, or trajectory-derived hindsight~\cite{ma2026sdsearchonpolicyhindsightselfdistillation,liang2026searche1selfdistillationdrivesselfevolution,yang2026selfdistilledrlvr,lu2026selfdistilledagenticreinforcementlearning}. However, obtaining a dense teacher signal does not by itself determine what the teacher should know or how its guidance should affect outcome-based credit.

Search-QA training instances often contain annotated supporting evidence that the agent must discover for itself at inference time. Unlike a golden answer, which primarily specifies the terminal target, supporting evidence identifies the information that a useful query should uncover without prescribing a single reference query. We therefore treat supporting evidence as \emph{instance-level privileged information}. This perspective exposes two coupled design decisions: \emph{\ding{172}~what privileged information should guide the sampled actions, and \ding{173}~how should that guidance be translated into credit---where in the trajectory should it act, and how should it interact with the outcome-derived advantage?}

We propose \textbf{EviSD}, an evidence-conditioned OPSD framework that addresses these two decisions with the same model acting as student and privileged teacher. The student samples each trajectory from the original inference-time context. During training only, the teacher re-scores the sampled actions with instance-specific privileged information, producing a detached token-level teacher--student gap.

\paragraph{\ding{172} What: Instance-Level Evidence as Privileged Information.}
For generated search actions, EviSD conditions the teacher on supporting evidence from the current QA instance. This evidence comes directly from training metadata, provides an instance-specific view of what a useful query should retrieve, and is not directly provided to the agent at inference. It is therefore the primary privileged signal for search in EviSD. For terminal answer actions, golden answers provide complementary target-side context.

\paragraph{\ding{173} How: Action-Localized, Outcome-Anchored Credit.}
EviSD separates where privileged guidance acts from how it enters optimization. \emph{Where:} the teacher--student gap affects only the content tokens of generated search and answer actions; generated non-action tokens retain their original GRPO credit, while retrieved passages remain masked environment observations. \emph{How:} EviSD maps the detached gap through a bounded function and uses it to modulate the magnitude of the outcome-derived advantage rather than optimizing a separate distillation objective. The outcome-derived GRPO advantage remains the anchor that determines the update direction, while privileged information refines how strongly individual action tokens inherit that credit. For $0<\lambda<1$, the correction cannot reverse the sign of the original advantage. Privileged context and teacher scoring are removed at inference, so the deployed agent is unchanged.

Our main contributions are as follows:
\begin{itemize}
\item We propose \textbf{EviSD}, which uses instance-level supporting evidence as privileged teacher context for search actions and golden answers as complementary context for terminal answer actions.
\item We introduce an action-localized, outcome-anchored credit modulation rule that converts the detached teacher--student gap into a bounded correction to GRPO advantages. It requires no auxiliary distillation loss or inference-time privileged information.
\item Across seven QA benchmarks and three backbones, EviSD exceeds the strongest compared method by 1.3--2.3 macro-average EM points. Controlled ablations show that answer-only context and full-response modulation reduce average EM by 2.1 and 7.1 points, respectively.
\end{itemize}

\section{Related Work}

\paragraph{Credit Assignment for Search Agents.}
Search-R1 and KBQA-R1 apply outcome-based RL to document search and multi-turn knowledge-base interaction, respectively~\cite{jin2025searchr1trainingllmsreason,sun2026kbqar1}. Because terminal rewards do not reveal which intermediate decisions produced the outcome, subsequent work seeks more localized credit signals. One line of research introduces finer-grained process signals. AutoRefine evaluates retrieval quality, IGPO and TIPS quantify changes in answer confidence, PiCA models cumulative progress, and StepSearch and CriticSearch rely on explicit process evaluation~\cite{shi2025searchrefinethinkfacilitating,wang2026informationgainbasedpolicyoptimization,xie2026tipsturnlevelinformationpotentialreward,liu2026picapivotbasedcreditassignment,wang2025stepsearchignitingllmssearch,zhang2025criticsearchfinegrainedcreditassignment}. A complementary line improves credit estimation through rollout structure. GiGPO and HGPO compare transitions from compatible states, while GraphGPO organizes rollout transitions into a graph for advantage aggregation~\cite{feng2025groupingrouppolicyoptimizationllm,he2026hierarchyofgroupspolicyoptimizationlonghorizon,cheng2026graphbasedcreditassignment}. Despite their different formulations, these methods all localize credit by introducing additional rewards, evaluators, or structural relations among rollout states.

\paragraph{Privileged On-Policy Self-Distillation.}
Privileged on-policy self-distillation offers a different source of supervision. OPSD evaluates the same on-policy response under the student context and a privileged teacher context, then distills the resulting token-level distributional difference~\cite{zhao2026selfdistilledreasoneronpolicyselfdistillation}. Related methods differ in how they construct privileged teacher contexts and incorporate the resulting signals into optimization. SD-Search derives hindsight supervision from grouped trajectories, Search-E1 uses an efficient sibling trajectory, Skill-SD extracts reusable skills from successful rollouts, and GAPD conditions a stop-gradient teacher on state-aligned gold actions for structured KBQA~\cite{ma2026sdsearchonpolicyhindsightselfdistillation,liang2026searche1selfdistillationdrivesselfevolution,wang2026skillsdskillconditionedselfdistillationmultiturn,sun2026gapd}. Selective hindsight distillation further studies which feedback sources should be used and where they should be inserted~\cite{li2026distillselectivehindsightdistillation}. On the optimization side, RLSD uses teacher and student divergence to rescale reward-based advantages and SDAR gates an auxiliary distillation objective~\cite{yang2026selfdistilledrlvr,lu2026selfdistilledagenticreinforcementlearning}.

EviSD differs from these methods in both the source and use of privileged information. Unlike trajectory-derived hindsight, task-level skills, or state-aligned gold actions, it provides instance-specific evidence to search actions and golden answers only to answer actions. Rather than optimizing an auxiliary distillation objective, EviSD uses the detached teacher--student gap only as a bounded modulation of outcome-derived action credit. This preserves the outcome objective while refining credit where the privileged information is most relevant.

\section{Method}

\begin{figure*}[t]
\centering
\includegraphics[width=\linewidth]{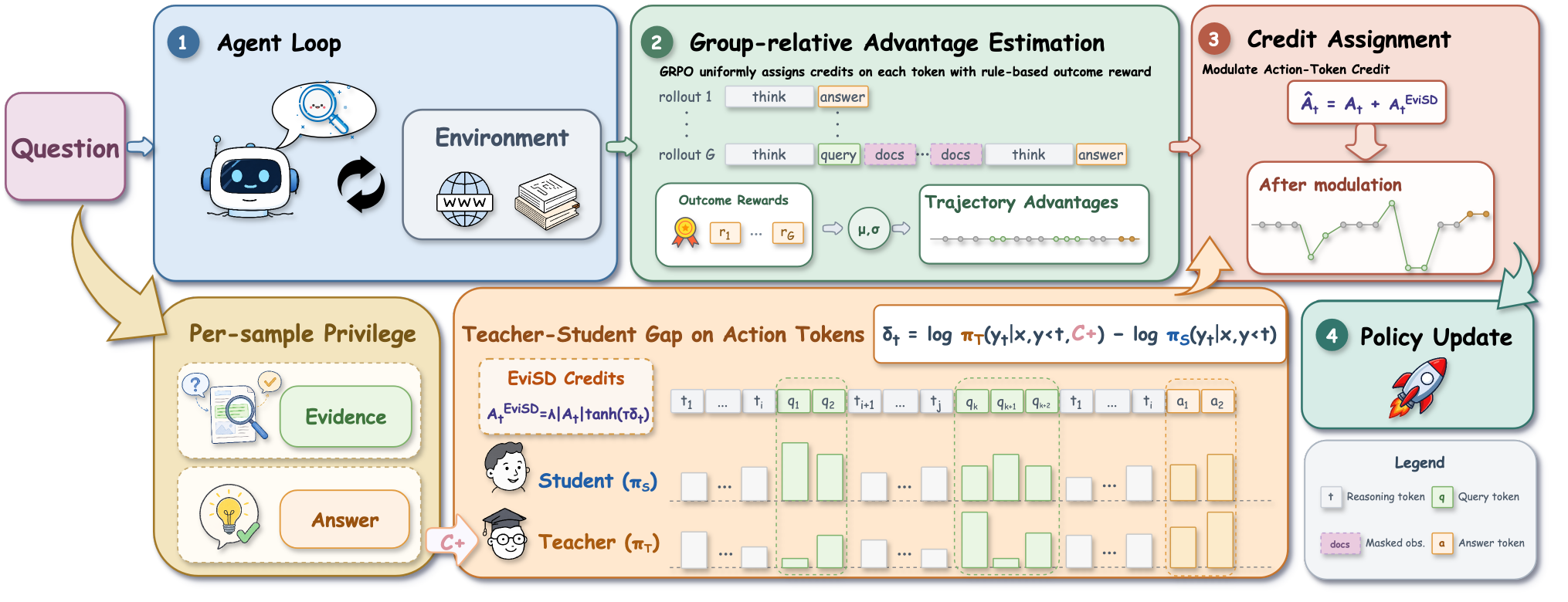}
\caption{Training framework of EviSD. Action-aligned privileged contexts produce self-distillation signals for bounded modulation of outcome-derived GRPO credits on search and answer tokens.}
\label{fig:framework}
\end{figure*}

We propose \textbf{EviSD}, an evidence-conditioned self-distillation framework that uses instance-level privileged information to refine outcome-derived credit on search and answer actions.
The method follows the two design decisions introduced above. It first constructs an instance-level privileged context aligned with each generated action. It then obtains a detached teacher--student contrast and translates that guidance into credit through two coordinated choices: localization determines \emph{where} the guidance acts, and outcome-anchored modulation determines \emph{how} it enters optimization.

\subsection{Problem Setup}

A search-augmented language model answers a question $q$ by interleaving generation with external retrieval. At turn $k$ of trajectory $i$, its context $x_{i,k}$ contains $q$, all preceding generated turns, and the passages returned by earlier searches. The policy generates a response
$y_{i,k}=(y_{i,k,1},\ldots,y_{i,k,T_{i,k}})$, which is parsed as an action $a_{i,k}\in\{\mathtt{search},\mathtt{answer},\mathtt{other}\}$. A search action sends its query to the retriever and appends the returned passages to the next-turn context; an answer action terminates the trajectory. Retrieved passages are environment observations rather than policy outputs, so only tokens generated by the policy participate in optimization.

During training, the old policy $\pi_{\theta_{\mathrm{old}}}$ samples a group of $G$ trajectories for the same question. Comparing each terminal answer against the golden answers yields an outcome reward $r_i$. GRPO standardizes these rewards within the group to obtain
\[
A_i =
\frac{r_i-\mu_r}{\sigma_r+\epsilon},
\qquad
\mu_r=\frac{1}{G}\sum_{j=1}^{G}r_j,
\]
where $\sigma_r$ is the group reward standard deviation. Then the trajectory advantage is broadcast to every generated response token:
\[
A_{i,k,t}=A_i m^{\mathrm{resp}}_{i,k,t},
\]
where $m^{\mathrm{resp}}_{i,k,t}=1$ on policy-generated response tokens and $0$ on prompts and retrieved observations. EviSD retains this outcome-derived advantage as the anchor and selectively refines it using privileged information available for the training instance.

\subsection{Instance-Level Privileged Context Construction}

For each training question, let $\mathcal{E}_q$ denote its annotated supporting evidence, grouped by source document, and let $\mathcal{C}_q$ denote its golden answer including aliases. The evidence identifies the critical information that a useful search should retrieve without prescribing a specific reference query, allowing the teacher to evaluate the student's own search strategy. The golden answers provide the corresponding instance-level privilege for the answer action. Both sources are used only to construct a teacher privileged context during training.

EviSD aligns the privilege source with the student's output action:
\[
c^{\mathrm{priv}}_{i,k} =
\left\{
\begin{array}{ll}
\mathcal{E}_q, & a_{i,k}=\mathtt{search},\ \mathcal{E}_q\neq\emptyset,\\
\mathcal{C}_q, & a_{i,k}=\mathtt{answer},\\
\emptyset, & \mathrm{otherwise}.
\end{array}
\right.
\]
For each sampled turn, EviSD leaves the interaction history unchanged and prepends a short, action-specific hint to construct the teacher privileged context
\[
\widetilde{x}_{i,k} = P(a_{i,k},c^{\mathrm{priv}}_{i,k}) \Vert x_{i,k},
\]
where $P$ formats supporting evidence for a search action and golden answers for an answer action. Thus, the teacher model can evaluate search actions with access to evidence about what the query should uncover, and answer actions with access to the golden answers. For invalid or unsupported actions, $c^{\mathrm{priv}}_{i,k}=\emptyset$ and $\widetilde{x}_{i,k}=x_{i,k}$.

\subsection{Translating Privileged Guidance into Credit}

\paragraph{Teacher--Student Contrast.}
The same policy provides the student and teacher views under different contexts. The student first samples $y_{i,k}$ on-policy from the original context $x_{i,k}$. The privileged teacher re-scores this exact response under $\widetilde{x}_{i,k}$.

We define the token-level self-distillation signal as the detached log-likelihood ratio
\[
\begin{array}{r@{}l}
\delta_{i,k,t}=\mathrm{sg}\!\big[&
\log\pi_\theta(y_{i,k,t}\mid\widetilde{x}_{i,k},y_{i,k,<t})\\[-1pt]
&-\log\pi_\theta(y_{i,k,t}\mid x_{i,k},y_{i,k,<t})\big].
\end{array}
\]
where $\mathrm{sg}$ stops gradients through both scoring views. A positive $\delta_{i,k,t}$ means that the action token is more compatible with the evidence- or answer-conditioned teacher, while a negative value indicates disagreement with that privileged view. 

\paragraph{Where: Action-Span Localization.}
The binary action mask $m^{\mathrm{act}}_{i,k,t}$ is defined over response tokens and equals one only for the content tokens of generated search and answer actions. Reasoning tokens, delimiters, retrieved text, and all other response tokens remain unmodified. The mask therefore localizes privileged guidance without changing which tokens participate in the underlying GRPO objective.

\paragraph{How: Outcome-Anchored Modulation.}
EviSD maps the distillation signal through the bounded modulation function
\[
g_\tau(\delta)=\tanh(\tau\delta),
\]
where $\tau$ controls how sensitively credit responds to the teacher--student contrast. The resulting \emph{privileged credit modulation term} is
\[
\begin{array}{r@{}l}
\Delta A^{\mathrm{priv}}_{i,k,t} ={}&
\lambda |A_{i,k,t}|
g_\tau(\delta_{i,k,t})\,m^{\mathrm{act}}_{i,k,t},\\
\widehat{A}_{i,k,t} ={}&
A_{i,k,t}+\Delta A^{\mathrm{priv}}_{i,k,t},
\end{array}
\]
where $\lambda$ controls the maximum relative correction.

The modulation is tied to the magnitude of the outcome advantage and is explicitly bounded:
\[
|\Delta A^{\mathrm{priv}}_{i,k,t}|
\leq \lambda |A_{i,k,t}|.
\]
Consequently, privileged information can adjust the strength of credit but cannot generate an update when the outcome-derived advantage is zero. GRPO determines the update direction from the trajectory outcome, while the privileged signal modulates its magnitude for each action token. When the privileged context favors a token, positive credit is strengthened and negative credit is softened. When it disfavors the token, the reverse occurs. The modulated credit therefore remains aligned with the update direction determined by the original GRPO advantage. For $\lambda<1$, $\widehat{A}_{i,k,t}$ preserves the sign of $A_{i,k,t}$, while tokens outside the action mask retain their original credits.

\subsection{Training Objective and Inference}

EviSD substitutes $\widehat{A}_{i,k,t}$ for the broadcast advantage in the standard clipped GRPO objective:
\[
\begin{array}{r@{}l}
J_{\mathrm{EviSD}}(\theta)=
\mathrm{E}_{i,k,t}\!\big[&
\min\!\left(\rho_{i,k,t}\widehat A_{i,k,t},
\bar\rho_{i,k,t}\widehat A_{i,k,t}\right)\\[-1pt]
&-\beta D_{\mathrm{KL}}(\pi_\theta\|\pi_{\mathrm{ref}})\big],
\end{array}
\]
where $\rho_{i,k,t}=\frac{\pi_\theta(y_{i,k,t}\mid x_{i,k},y_{i,k,<t})}{\pi_{\theta_{\mathrm{old}}}(y_{i,k,t}\mid x_{i,k},y_{i,k,<t})}$ and $\bar\rho_{i,k,t}=\mathrm{clip}(\rho_{i,k,t},1-\eta,1+\eta)$. The expectation averages optimized response tokens, and the routed privilege affects optimization only through the $\widehat A_{i,k,t}$.

Each update follows the original on-policy interaction loop. EviSD first samples and executes a group of trajectories, obtains their outcome rewards, and computes $A_i$. It then constructs the appropriate privileged context for every generated search or answer turn, re-scores the sampled action once, and forms $\widehat A_{i,k,t}$ before applying the standard GRPO update. The additional teacher view is needed only for training. At inference time, the original policy interacts with the retriever without privileged context.

\begin{table*}[t]
\centering
\setlength{\tabcolsep}{5pt}
\begin{tabular}{l ccc cccc c}
\toprule
\multirow{2}{*}{\textbf{Method}} & \multicolumn{3}{c}{\textbf{Single-hop}} & \multicolumn{4}{c}{\textbf{Multi-hop}} & \multirow{2}{*}{\textbf{Avg.}} \\
\cmidrule(lr){2-4} \cmidrule(lr){5-8}
 & NQ & TriviaQA & PopQA & HotpotQA & 2WikiMQA & MuSiQue & Bamboogle & \\
\midrule
\multicolumn{9}{l}{\textit{Outcome-Based RL and Reward Shaping}} \\
Search-R1~\cite{jin2025searchr1trainingllmsreason} & 42.9 & 62.3 & 42.7 & 38.6 & 34.6 & 16.2 & 40.0 & 39.6 \\
AutoRefine~\cite{shi2025searchrefinethinkfacilitating} & 42.0 & 60.4 & 41.3 & 39.8 & 33.3 & 17.6 & 39.2 & 39.1 \\
\midrule
\multicolumn{9}{l}{\textit{Turn-/Step-Level Credit Assignment}} \\
IGPO~\cite{wang2026informationgainbasedpolicyoptimization} & 26.5 & 55.2 & 40.5 & 36.1 & 27.7 & 8.6 & 35.2 & 32.8 \\
TIPS~\cite{xie2026tipsturnlevelinformationpotentialreward} & 43.4 & 64.3 & 44.5 & 43.0 & 43.0 & 17.1 & 36.8 & 41.7\\
PiCA~\cite{liu2026picapivotbasedcreditassignment} & 46.0 & 64.1 & 44.2 & 42.4 & 40.1 & 19.7 & 41.9 & 42.6\\
StepSearch$^\diamondsuit$~\cite{wang2025stepsearchignitingllmssearch} & - & - & - & 38.6 & 36.6 & \underline{22.6} & 40.0 & -\\
CriticSearch$^\diamondsuit$~\cite{zhang2025criticsearchfinegrainedcreditassignment} & - & - & - & 44.2 & 42.8 & 19.4 & 47.2 & -\\
GiGPO~\cite{feng2025groupingrouppolicyoptimizationllm} & 46.4 & 64.7 & 46.1 & 41.6 & 43.6 & 18.9 & 68.9 & 47.2 \\
\midrule
\multicolumn{9}{l}{\textit{Self-Distillation: Rollout-Conditioned Privilege}} \\
SD-Search~\cite{ma2026sdsearchonpolicyhindsightselfdistillation} & \textbf{49.5} & \underline{66.8} & \underline{49.4} & \underline{47.1} & 44.1 & 22.2 & 54.4 & 47.6 \\
Search-E1~\cite{liang2026searche1selfdistillationdrivesselfevolution} & 48.7 & \textbf{69.4} & \textbf{49.7} & 46.9 & 45.5 & \textbf{23.6} & 53.4 & 48.2 \\
\midrule
\multicolumn{9}{l}{\textit{Self-Distillation: Skill-Conditioned Privilege}} \\
OPSD$_{Skill}$~\cite{zhao2026selfdistilledreasoneronpolicyselfdistillation} & 8.8 & 17.5 & 2.5 & 8.6 & 4.2 & 0.5 & 1.2 & 6.2 \\
Skill-SD~\cite{wang2026skillsdskillconditionedselfdistillationmultiturn} & 47.1 & 47.8 & 44.2 & \textbf{64.5} & 42.1 & 20.2 & 69.0 & 47.8 \\
RLSD$_{Skill}$~\cite{yang2026selfdistilledrlvr} & 46.8 & 63.0 & 44.4 & 45.5 & \textbf{48.9} & 21.5 & \underline{73.0} & \underline{49.0} \\
SDAR~\cite{lu2026selfdistilledagenticreinforcementlearning} & 46.3 & 63.5 & 48.2 & 43.8 & 48.4 & 19.6 & \underline{73.0} & \underline{49.0} \\
\midrule
\multicolumn{9}{l}{\textit{Self-Distillation: Answer-Conditioned Privilege}} \\
OPSD$_{Ans}$~\cite{zhao2026selfdistilledreasoneronpolicyselfdistillation} & 6.2 & 17.3 & 9.8 & 8.8 & 14.8 & 0.9 & 1.2 & 8.4 \\
RLSD$_{Ans}$~\cite{yang2026selfdistilledrlvr} & 46.7 & 64.2 & 45.5 & 41.8 & 42.9 & 16.6 & 65.7 & 46.2 \\
\midrule
\multicolumn{9}{l}{\textit{Self-Distillation: Instance-Level Privilege}} \\
\rowcolor{gray!10}
\textbf{EviSD (Ours)} & \underline{48.9} & 66.5 & 48.6 & 46.6 & \underline{48.6} & 21.6 & \textbf{74.6} & \textbf{50.8} \\
\bottomrule
\end{tabular}
\caption{Main results (EM) with Qwen2.5-7B-Instruct. Best results are \textbf{bold}; second-best results are \underline{underlined}. $^\diamondsuit$ indicates that StepSearch and CriticSearch are trained on only one multi-hop dataset.}
\label{tab:main_results}
\end{table*}

\section{Experiments}

\subsection{Experimental Setup}

\paragraph{Datasets.}
We evaluate EviSD on seven search-augmented QA benchmarks, including three single-hop datasets, NQ~\cite{kwiatkowski-etal-2019-natural}, TriviaQA~\cite{joshi-etal-2017-triviaqa}, and PopQA~\cite{mallen-etal-2023-trust}, and four multi-hop datasets, HotpotQA~\cite{yang2018hotpotqadatasetdiverseexplainable}, 2WikiMultiHopQA~\cite[2WikiMQA;][]{ho2020constructingmultihopqadataset}, MuSiQue~\cite{trivedi2022musiquemultihopquestionssinglehop}, and Bamboogle~\cite{press2023measuringnarrowingcompositionalitygap}. NQ and HotpotQA are in-domain datasets, while the remaining five are used to evaluate out-of-domain generalization. We report Exact Match (EM) following prior works.

\paragraph{Baselines.}
We compare EviSD with three categories of methods: outcome-based RL and reward shaping, including Search-R1~\cite{jin2025searchr1trainingllmsreason} and AutoRefine~\cite{shi2025searchrefinethinkfacilitating}; turn- or step-level credit assignment, including IGPO~\cite{wang2026informationgainbasedpolicyoptimization}, TIPS~\cite{xie2026tipsturnlevelinformationpotentialreward}, PiCA~\cite{liu2026picapivotbasedcreditassignment}, StepSearch~\cite{wang2025stepsearchignitingllmssearch}, CriticSearch~\cite{zhang2025criticsearchfinegrainedcreditassignment}, and GiGPO~\cite{feng2025groupingrouppolicyoptimizationllm}; and on-policy self-distillation methods. We subdivide the last category by teacher context: rollout-conditioned privilege for SD-Search~\cite{ma2026sdsearchonpolicyhindsightselfdistillation} and Search-E1~\cite{liang2026searche1selfdistillationdrivesselfevolution}; skill-conditioned privilege for OPSD$_{Skill}$, Skill-SD, RLSD$_{Skill}$, and SDAR~\cite{zhao2026selfdistilledreasoneronpolicyselfdistillation,wang2026skillsdskillconditionedselfdistillationmultiturn,yang2026selfdistilledrlvr,lu2026selfdistilledagenticreinforcementlearning}; and answer-conditioned privilege for OPSD$_{Ans}$ and RLSD$_{Ans}$. Unless otherwise specified, all methods use the same training data and retrieval environment. For IGPO, we reproduce its 7B training results in our environment rather than using its originally reported results based on web search.

\paragraph{Implementation Details.}
We use Qwen2.5-7B-Instruct~\cite{qwen2025qwen25technicalreport} as the main base model. The retrieval environment follows Search-R1, using the 2018 Wikipedia dump~\cite{karpukhin-etal-2020-dense} and E5~\cite{wang2024textembeddingsweaklysupervisedcontrastive} as the dense retriever. Each query retrieves the top-3 passages, and an episode allows at most four searches. For search-action privilege, we group the dataset-annotated supporting sentences by source document and retain at most two documents and two sentences per document. For answer actions, we use the golden answers. In extension experiments, each relevant substitute document is constructed from a non-supporting paragraph in HotpotQA's \textit{distractor} field, retaining its first two sentences. We train with GRPO~\cite{shao2024deepseekmathpushinglimitsmathematical} for 300 steps on 8 GPUs, using 8 rollouts per question, 128 questions per step, a learning rate of $1\times10^{-6}$ with 10\% warmup, a KL coefficient of $0.001$, and a clip ratio of $0.2$. Rollouts use vLLM with temperature 1.0. Unless otherwise specified, EviSD uses $\lambda=0.2$ and $\tau=5$. We additionally evaluate Qwen2.5-3B-Instruct and Qwen3-1.7B to test robustness across model scales and generations. All reported metrics are obtained from a single evaluation run.

\subsection{Experimental Results}
We first report overall benchmark performance. We then organize the analysis around EviSD's two main design decisions. RQ1 examines instance-level evidence as privileged context. RQ2 studies how privileged guidance is translated into credit. RQ3 examines on-policy training dynamics, and RQ4 tests sensitivity to privileged-evidence quality.

\paragraph{Overall Performance.}

\begin{figure}[t]
\centering
\includegraphics[width=0.7\linewidth]{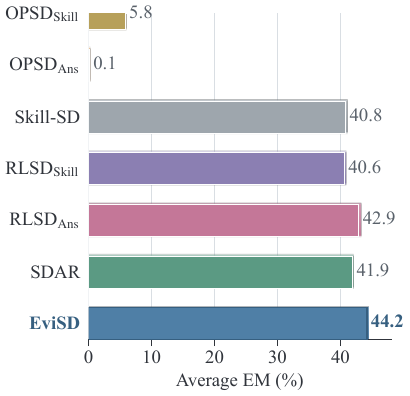}
\caption{Average EM of self-distillation methods with Qwen3-1.7B.}
\label{fig:qwen3_1_7b_average_em}
\end{figure}

\begin{table*}[t]
\centering
\setlength{\tabcolsep}{5pt}
\begin{tabular}{l ccc cccc c}
\toprule
\multirow{2}{*}{\textbf{Method}} & \multicolumn{3}{c}{\textbf{Single-hop}} & \multicolumn{4}{c}{\textbf{Multi-hop}} & \multirow{2}{*}{\textbf{Avg.}} \\
\cmidrule(lr){2-4} \cmidrule(lr){5-8}
 & NQ & TriviaQA & PopQA & HotpotQA & 2WikiMQA & MuSiQue & Bamboogle & \\
\midrule
Search-R1~\cite{jin2025searchr1trainingllmsreason} & 39.7 & 56.5 & 39.1 & 33.1 & 31.0 & 12.4 & 23.2 & 33.6 \\
AutoRefine~\cite{shi2025searchrefinethinkfacilitating} & 43.6 & 59.7 & 44.7 & 40.4 & 38.0 & 16.9 & 33.6 & 39.6 \\
\midrule
TIPS~\cite{xie2026tipsturnlevelinformationpotentialreward} & 43.5 & 58.8 & 42.8 & 31.4 & 29.3 & 8.7 & 20.8 & 33.6 \\
PiCA~\cite{liu2026picapivotbasedcreditassignment} & 42.6 & 61.2 & 41.7 & 40.0 & 40.8 & 16.0 & 34.7 & 39.6 \\
StepSearch$^\diamondsuit$~\cite{wang2025stepsearchignitingllmssearch} & -- & -- & -- & 34.5 & 32.0 & 17.4 & 34.4 & -- \\
CriticSearch$^\diamondsuit$~\cite{zhang2025criticsearchfinegrainedcreditassignment} & -- & -- & -- & 41.4 & 40.9 & 18.0 & 36.8 & -- \\
GiGPO~\cite{feng2025groupingrouppolicyoptimizationllm} & 42.0 & 59.5 & 42.4 & 36.9 & 37.0 & 12.6 & 64.1 & 42.1 \\
\midrule
SD-Search~\cite{ma2026sdsearchonpolicyhindsightselfdistillation} & \underline{46.9} & 62.4 & 46.5 & \underline{42.4} & 41.9 & \underline{18.4} & 40.4 & 42.7 \\
Search-E1~\cite{liang2026searche1selfdistillationdrivesselfevolution} & \textbf{47.4} & \underline{62.6} & 46.1 & \textbf{42.7} & \underline{43.6} & \textbf{19.3} & 46.4 & 44.0 \\
\midrule
OPSD$_{Skill}$~\cite{zhao2026selfdistilledreasoneronpolicyselfdistillation} & 0.1 & 0.1 & 0.1 & 0.0 & 0.0 & 0.0 & 0.0 & 0.0 \\
Skill-SD~\cite{wang2026skillsdskillconditionedselfdistillationmultiturn} & 44.4 & 60.4 & 44.0 & 39.5 & 40.4 & 15.4 & 64.9 & 44.1 \\
RLSD$_{Skill}$~\cite{yang2026selfdistilledrlvr} & 41.5 & 58.6 & 42.3 & 40.4 & 40.2 & 16.8 & \underline{66.9} & 43.8 \\
SDAR~\cite{lu2026selfdistilledagenticreinforcementlearning} & 44.8 & 58.1 & 44.3 & 38.6 & 36.2 & 15.7 & 66.1 & 43.4 \\
\midrule
OPSD$_{Ans}$~\cite{zhao2026selfdistilledreasoneronpolicyselfdistillation} & 0.0 & 0.0 & 0.0 & 0.0 & 0.0 & 0.0 & 0.0 & 0.0 \\
RLSD$_{Ans}$~\cite{yang2026selfdistilledrlvr} & 44.8 & 61.8 & \underline{47.5} & 40.1 & 41.3 & 14.6 & 64.1 & \underline{44.9} \\
\midrule
\rowcolor{gray!10}
\textbf{EviSD (Ours)} & 46.6 & \textbf{63.1} & \textbf{48.7} & \underline{42.4} & \textbf{44.9} & 17.5 & \textbf{67.3} & \textbf{47.2} \\
\bottomrule
\end{tabular}
\caption{Main results (EM) with Qwen2.5-3B-Instruct. Best results are \textbf{bold}; second-best results are \underline{underlined}. $^\diamondsuit$ indicates that StepSearch and CriticSearch are trained on only one multi-hop dataset.}
\label{tab:small_model_results}
\end{table*}
As shown in Tables~\ref{tab:main_results} and~\ref{tab:small_model_results} and Figure~\ref{fig:qwen3_1_7b_average_em}, EviSD achieves the highest average EM across all evaluated settings, consistently outperforming the previous state-of-the-art method by 1.3--2.3 points. These consistent gains across Qwen2.5-3B, Qwen2.5-7B, and Qwen3-1.7B demonstrate the robustness and generalizability of our method across model scales and generations.
The 7B search-call diagnostic further shows that EviSD reaches higher success with only 1.87 searches per trajectory, compared with 2.16 for SDAR and 2.27 for RLSD. This reduction indicates more precise guidance of search actions and fewer redundant retrievals. Additional training diagnostics and hyperparameter analyses are provided in the supplementary material.

\paragraph{RQ1: What privileged context is most useful for supervising search actions?}
As shown in Tables~\ref{tab:main_results} and~\ref{tab:small_model_results} and Figure~\ref{fig:qwen3_1_7b_average_em}, EviSD outperforms the strongest skill-conditioned baseline by 1.8--3.1 average EM points and the strongest answer-conditioned baseline by 1.3--4.6 points. Because these methods also differ in how distillation is integrated into optimization, the cross-method comparison does not isolate the privilege source. We therefore conduct a controlled ablation under the same EviSD training setup. Replacing search evidence with answer-only privileged context reduces average EM from 50.8 to 48.7, as shown in Table~\ref{tab:ablation}. This controlled result directly supports instance-specific evidence over answer-only context for supervising search actions; the cross-method results are consistent with, but do not independently isolate, its advantage over retrieved skills. Retrieved skills provide reusable guidance but do not directly identify the target evidence for the current question, whereas golden answers provide only limited guidance on what each intermediate query should retrieve.

\begin{table}[ht]
\centering
\small
\setlength{\tabcolsep}{2.4pt}
\begin{tabular}{@{}lrrrr@{}}
\toprule
\textbf{Variant} & \textbf{S-hop} & \textbf{M-hop} & \textbf{All} & $\Delta$ \\
\midrule
\rowcolor{gray!10}
\textbf{Full EviSD (ours)} & \textbf{54.7} & \underline{47.9} & \textbf{50.8} & -- \\
\midrule
\multicolumn{5}{@{}l}{\textit{Privileged Context Construction}} \\
\quad Evidence + answer for all actions & 54.0 & \textbf{48.0} & \underline{50.6} & $-0.2$ \\
\quad Answer for all actions & 52.1 & 46.1 & 48.7 & $-2.1$ \\
\midrule
\multicolumn{5}{@{}l}{\textit{Translating Privileged Guidance into Credit}} \\
\quad Full-response modulation & 49.7 & 39.2 & 43.7 & $-7.1$ \\
\quad Response-wide auxiliary loss & \underline{54.1} & 38.1 & 45.0 & $-5.8$ \\
\bottomrule
\end{tabular}
\caption{Component ablations of EviSD with Qwen2.5-7B-Instruct. S-hop, M-hop, and All denote macro-averaged EM; $\Delta$ is relative to full EviSD. Best and second-best results are \textbf{bold} and \underline{underlined}.}
\label{tab:ablation}
\end{table}

\begin{figure}[t]
\centering
\includegraphics[width=0.85\linewidth]{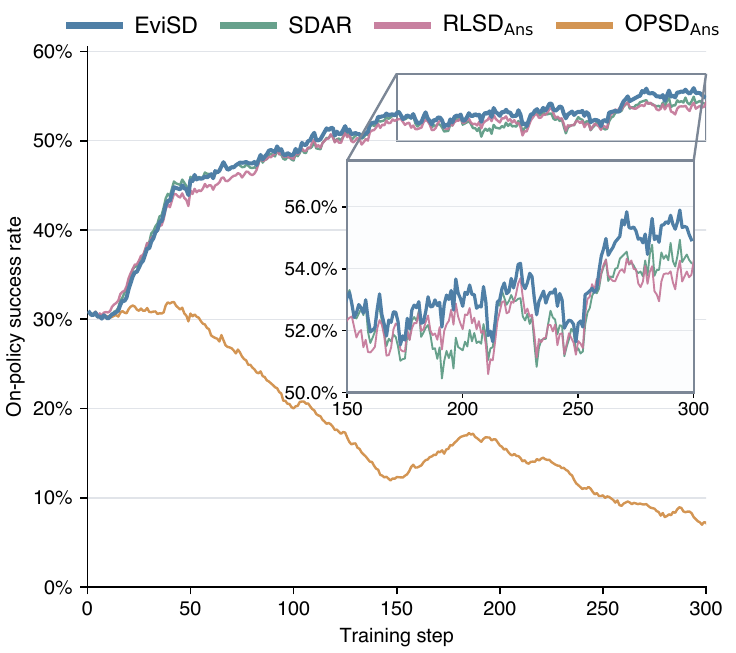}
\caption{On-policy success rate during Qwen2.5-7B training.}
\label{fig:episode_success_rate}
\end{figure}

Table~\ref{tab:ablation} also examines how privilege sources are assigned across actions. Sharing the combined evidence--answer context across actions reaches 50.6 average EM, only 0.2 points below action-conditioned assignment, whereas answer-only context lowers it to 48.7. Thus, evidence availability is the critical driver of performance gains, whereas strict routing primarily provides semantic alignment with negligible performance degradation.

\paragraph{RQ2: How does EviSD translate privileged guidance into credit?}
Removing action-span localization while retaining bounded modulation lowers average EM by 7.1 points, directly supporting the concentration of privileged guidance on generated action content. The response-wide SDAR-style auxiliary loss retains the same action-conditioned teacher contexts but applies gated distillation to all valid response tokens instead of modulating action-token credit. It performs 5.8 points below EviSD, with the largest degradation on multi-hop tasks. Because this alternative changes both the optimization objective and token scope, it tests whether dense privileged distillation can replace the complete EviSD credit design rather than isolating the loss form alone.
Across complete training trajectories and all three model scales, only 6.7\%--15.1\% of response tokens receive modulated credit. For teacher-compatible action tokens, the detached gap strengthens positive credit or softens negative credit, with the reverse effect for teacher-disfavored tokens. All other response tokens retain their original GRPO credit. This localization avoids response-wide privileged supervision and direct imitation of a teacher conditioned on information unavailable at inference.

\paragraph{RQ3: What training dynamics accompany evidence-conditioned credit modulation?}
Figure~\ref{fig:episode_success_rate} shows that EviSD sustains strong on-policy performance throughout training. Its smoothed curve leads during rapid first-half improvement and remains above the competing methods after step 150. Its mean success rate across second-half checkpoints reaches 53.6\%, exceeding the strongest competitor by 0.9 percentage points. This sustained separation is consistent with evidence-conditioned modulation continuing to provide useful credit as the policy evolves.

\begin{figure}[t]
\centering
\includegraphics[width=0.8\linewidth]{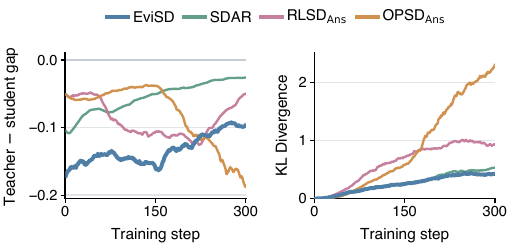}
\caption{Teacher--student gap and KL divergence during Qwen2.5-7B training.}
\label{fig:teacher_gap_kl}
\end{figure}
We further examine the teacher--student gap, which captures privileged contrast, and KL divergence, which measures deviation from the reference model. Figure~\ref{fig:teacher_gap_kl} shows that EviSD's gap remains separated from zero while its KL increases smoothly and stays at or near the lowest level among the compared methods. Together with the success curve, this pattern is consistent with gradual policy improvement without abrupt reference-policy deviation. In contrast, SDAR's gap approaches zero, RLSD's KL rises faster, and OPSD exhibits sharp increases in both gap magnitude and KL as its success collapses.

% \begin{table}[!ht]
% \centering
% \small
% \setlength{\tabcolsep}{5pt}
% \begin{tabular}{@{}lc@{}}
% \toprule
% \textbf{Search-action teacher context} & \textbf{Avg.} \\
% \midrule
% \rowcolor{gray!10}
% EviSD (2 gold docs.) & \textbf{44.2} \\
% \quad w/ 1 gold doc. + 1 distractor doc. & \textbf{44.2} \\
% \quad w/ 2 distractor docs. & \underline{43.6} \\
% % \quad w/ top-2 retrieved docs. & 42.3 \\
% \bottomrule
% \end{tabular}
% \caption{Sensitivity to privileged-context quality on Qwen3-1.7B-Instruct.}
% \label{tab:evidence_quality}
% \end{table}

% \paragraph{RQ4: How sensitive is EviSD to the quality of privileged evidence?}
% Table~\ref{tab:evidence_quality} varies the evidence supplied to the privileged teacher during training while leaving inference unchanged. On Qwen3-1.7B, replacing one of the two gold sources with a relevant non-supporting document from HotpotQA's \textit{distractor} field leaves the observed average EM unchanged at 44.2; replacing both reduces it by only 0.6 points to 43.6. EviSD is therefore robust to partial degradation of privileged evidence in this setting, although these results do not establish arbitrary documents as a general replacement for annotated evidence.

\begin{table}[!ht]
\centering
\small
\setlength{\tabcolsep}{5pt}
\begin{tabular}{@{}lc@{}}
\toprule
\textbf{Search-action teacher context} & \textbf{Avg.} \\
\midrule
\rowcolor{gray!10}
EviSD (2 gold docs.) & \textbf{44.2} \\
\quad w/ 1 gold docs.+ 1 relevant doc. & \textbf{44.2} \\
\quad w/ 2 relevant docs. & 43.6 \\
% \quad w/ top-2 retrieved docs. & 42.3 \\
\bottomrule
\end{tabular}
\caption{Sensitivity of EviSD to privileged-evidence quality on Qwen3-1.7B.}
\label{tab:evidence_quality}
\end{table}
\paragraph{RQ4: How sensitive is EviSD to the quality of privileged evidence available during training?}
Table~\ref{tab:evidence_quality} evaluates EviSD on Qwen3-1.7B when supporting evidence is progressively replaced with relevant documents. Replacing one of the two gold sources leaves the average EM unchanged at 44.2, while replacing both reduces it by only 0.6 points to 43.6. These results further suggest that relevant retrieved documents may serve as a viable source of privileged supervision, providing useful guidance even when fully annotated gold evidence is unavailable.

\section{Conclusion}
We presented EviSD, an evidence-conditioned self-distillation framework for coarse credit assignment in multi-turn search agents. EviSD aligns supporting evidence with search actions and golden answers with answer actions, then converts the detached teacher--student gap into bounded, action-localized modulation of outcome-derived GRPO credit. Without inference-time changes, it achieves the best overall performance among the compared methods across seven QA benchmarks and three backbones. Controlled analyses further validate both design choices, show stable training dynamics, and indicate limited sensitivity to the tested relevant-document substitutions.

\bibliography{aaai2027}

\clearpage
\appendix

\section{Supplementary Material}

\subsection{Rollout Prompt and Privileged Teacher Scoring}

\paragraph{Rollout prompt.}
At each turn, the environment reconstructs the interaction as a single user
message and applies the backbone's chat template. The following prompt is used
for both training rollouts and evaluation:

\begin{center}
\setlength{\fboxsep}{5pt}
\fcolorbox{black!55}{black!3}{\begin{minipage}{0.91\columnwidth}
\scriptsize\ttfamily\raggedright
You are an expert agent tasked with answering the given question
step-by-step.\\
Your question: \{question\}\\[2pt]
\textnormal{\textit{[Inserted after the first turn]}}\\
Prior to this step, you have already taken \{step\_count\} step(s).
Below is the interaction history where <search> </search> wrapped your past
search queries and <documents> </documents> wrapped the corresponding search
results returned by the external search engine. History:\\
\{history\}\\[2pt]
Now it's your turn to respond for the current step.\\
You should first conduct reasoning process. This process MUST be enclosed
within <think> </think> tags.\\
After completing your reasoning, choose only one of the following actions
(do not perform both):\\
(1) If you find you lack some knowledge, you can call a search engine to get
more external information using format: <search> your query </search>.\\
(2) If you have enough knowledge to answer the question confidently, provide
your final answer within <answer> </answer> tags, without detailed
illustrations. For example, <answer>Beijing</answer>.
\end{minipage}}
\end{center}

The history block is omitted at the first turn. Thereafter,
\texttt{\{history\}} contains previous search actions and their retrieved
passages within \texttt{<documents>} tags. Each valid response contains a
reasoning span followed by exactly one search or answer action.

\paragraph{Privileged teacher input.}
The student samples an action once from the rollout prompt above. EviSD then
prepends an action-dependent field to the unchanged prompt and asks the same
policy to re-score that sampled action. The action is not regenerated under the
privileged context. Table~\ref{tab:teacher_input} gives the exact teacher-only
prefixes.

\begin{center}
\centering
\small
\setlength{\tabcolsep}{3pt}
\begin{tabular}{@{}>{\raggedright\arraybackslash}p{0.16\columnwidth}
                    >{\raggedright\arraybackslash}p{0.74\columnwidth}@{}}
\toprule
\textbf{Action} & \textbf{Prefix prepended to the rollout prompt} \\
\midrule
Search &
\texttt{[Privileged Hint]}\newline
\texttt{Golden Evidence:}\newline
\texttt{\{supporting evidence\}} \\
Answer &
\texttt{[Privileged Hint]}\newline
\texttt{Golden Answers:}\newline
\texttt{\{answer aliases\}} \\
Other & Empty \\
\bottomrule
\end{tabular}
\captionof{table}{Action-dependent prefixes used for privileged scoring.}
\label{tab:teacher_input}
\end{center}

Supporting evidence is serialized as \texttt{title: text} entries, retaining
at most two documents and two annotated sentences per document. Each entry is
truncated to 400 characters and the complete field to 1,000 characters.
Golden-answer aliases are separate bullet points. Without supporting evidence,
teacher and student contexts coincide and the resulting gap is zero.

\paragraph{Action-span mask.}
The mask is constructed after decoding the sampled response. Only tokens
overlapping the content of a complete search or answer tag pair are selected:

\begin{center}
\setlength{\fboxsep}{2pt}
\fbox{\begin{minipage}{0.91\columnwidth}
\small\centering
\texttt{<think> reasoning tokens </think>}\\[3pt]
\texttt{<search>}\,
\colorbox{black!12}{\strut\texttt{search query}}\,
\texttt{</search>}\\[4pt]
\textit{or}\\[-1pt]
\texttt{<think> reasoning tokens </think>}\\[3pt]
\texttt{<answer>}\,
\colorbox{black!12}{\strut\texttt{answer tokens}}\,
\texttt{</answer>}
\end{minipage}}
\end{center}

A token is
selected whenever its decoded offset overlaps this span, including subwords
with a leading space. Reasoning tokens and incomplete tag pairs are excluded.

\subsection{Case Studies of Evidence-Conditioned Credit Modulation}

\paragraph{Within-rollout discrimination under an incorrect outcome.}
For case studies, we select a random subset of the Qwen2.5-7B-Instruct
rollouts collected at training step 300. Each question has one
instance-specific evidence set shared by all of its sampled rollouts. The
first case examines a single incorrect-outcome rollout whose search combines
an evidence-supported entity with a logically inconsistent title. Boldface
marks each analyzed key token and its matching occurrence in the golden
evidence when one exists.

\begin{center}
\setlength{\fboxsep}{4pt}
\fcolorbox{black!55}{black!3}{\begin{minipage}{0.90\columnwidth}
\footnotesize\raggedright
\textbf{Golden evidence for this question}\\[-1pt]
\textit{Steffan Rhodri:} \ldots{} portrayed \textbf{Reg} Cattermole in
``Harry Potter and the Deathly Hallows'' part I.\\
\textit{Harry Potter and the Deathly Hallows:} The final novel was published
by Bloomsbury in the United Kingdom \ldots{}\\[4pt]
\hrule
\vspace{3pt}
\textbf{Incorrect-outcome rollout}\\[-1pt]
\texttt{<think> \ldots{} The search results indicate that Reg Trotter is a role
in ``Rock and Chips.'' \ldots{} </think>}\\
\texttt{<search>}\textbf{\texttt{Reg}}\texttt{ Cattermole in the movie
}\textbf{\texttt{Rock}}\texttt{ and Chips</search>}
\end{minipage}}
\end{center}

\begin{center}
\footnotesize
\setlength{\tabcolsep}{3.5pt}
\begin{tabular}{@{}llrrrr@{}}
\toprule
\textbf{Relation} & \textbf{Key token} & $A$ & $\delta$ &
$\Delta A^{\mathrm{priv}}$ & $\widehat A$ \\
\midrule
Aligned & \texttt{Reg} & $-0.372$ & $0.116$ & $+0.039$ & $-0.333$ \\
Misaligned & \texttt{Rock} & $-0.372$ & $-11.628$ & $-0.074$ & $-0.446$ \\
\bottomrule
\end{tabular}
\captionof{table}{Token-level credit within one incorrect-outcome rollout.}
\label{tab:within_rollout_credit_case}
\end{center}

Every action token begins with the same outcome-derived advantage
$A=-0.372$. Yet the privileged context produces sharply different
teacher--student gaps for the two key tokens. It slightly favors the
evidence-supported \texttt{Reg} with $\delta=0.116$, while strongly rejecting
\texttt{Rock} with $\delta=-11.628$, a contrast of $11.744$ log-probability
units. This contrast pinpoints the unsupported association of Reg Cattermole
with \emph{Rock and Chips} in the search query. Under the bounded modulation,
the extreme negative gap for
\texttt{Rock} saturates $g_\tau$ and reaches the permitted correction
$-\lambda|A|\approx-0.074$. The moderate positive gap for \texttt{Reg} yields
a proportional correction of $+0.039$. Its credit is therefore softened to
$-0.333$, while the penalty on \texttt{Rock} is strengthened to $-0.446$.
The gap remains highly discriminative without inducing an unbounded update.

\paragraph{Across-rollout refinement under distinct outcomes.}
We next compare two searches at the same turn position for a question asking
which remake of \emph{Sethu} featured Darshan Kumar. One rollout yields a
correct outcome and the other an incorrect outcome. Both searches are re-scored
with the same golden evidence:

\begin{center}
\setlength{\fboxsep}{4pt}
\fcolorbox{black!55}{black!3}{\begin{minipage}{0.90\columnwidth}
\footnotesize\raggedright
\textbf{Shared golden evidence}\\[-1pt]
\textit{Darshan Kumar:} \ldots{} appeared in the 2003 Salman Khan starrer
\textbf{movie} ``Tere Naam'' \ldots\\
\textit{Tere Naam:} Tere Naam is a 2003 Indian Hindi-language romantic drama
film. The film is a remake of Bala's Tamil blockbuster ``Sethu'' (1999).
\end{minipage}}
\end{center}

\begin{center}
\setlength{\fboxsep}{4pt}
\fcolorbox{black!55}{black!3}{\begin{minipage}{0.90\columnwidth}
\footnotesize\ttfamily\raggedright
\textbf{Correct-outcome rollout}\\[-1pt]
<think> \ldots{} The primary remake of ``Sethu'' is in Hindi as ``Tere Naam,''
\ldots{} I need to explore if Darshan Kumar starred in any known remakes of
``Sethu.'' </think>\\
<search>Darshan Kumar in the \textbf{movie} Tere Naam</search>\\[3pt]
\hrule
\vspace{3pt}
\textbf{Incorrect-outcome rollout}\\[-1pt]
<think> \ldots{} Darshan Kumar predominantly works in the Kannada film
industry \ldots{} let's check if Darshan Kumar stars in any other adaptations
of ``Sethu.'' </think>\\
<search>Darshan Kumar in the \textbf{Kann}ada remake of Sethu</search>
\end{minipage}}
\end{center}

\begin{center}
\footnotesize
\setlength{\tabcolsep}{3.5pt}
\begin{tabular}{@{}llrrrr@{}}
\toprule
\textbf{Rollout} & \textbf{Key token} & $A$ & $\delta$ &
$\Delta A^{\mathrm{priv}}$ & $\widehat A$ \\
\midrule
Correct & \texttt{movie} & $0.814$ & $2.223$ & $+0.163$ & $0.976$ \\
Incorrect & \texttt{Kann} & $-1.183$ & $-0.653$ & $-0.236$ & $-1.420$ \\
\bottomrule
\end{tabular}
\captionof{table}{Step-300 search credit under distinct rollout outcomes.}
\label{tab:search_credit_case}
\end{center}

The token-level gaps again carry a sharp semantic contrast. The privileged
teacher favors the evidence-supported \texttt{movie} with $\delta=2.223$ but
rejects the contradictory \texttt{Kann} with $\delta=-0.653$, a difference of
$2.876$ log-probability units. At $\tau=5$, both gaps drive $g_\tau$ close to
its respective bound. Their corrections consequently approach the
outcome-scaled limits, reaching $+0.163$ for \texttt{movie} and $-0.236$ for
\texttt{Kann}. GRPO supplies the original positive and negative update
directions, while the privileged gaps sharpen their token-specific strengths.
The distance between the two credits increases from $1.997$ to $2.396$ without
changing either sign. Together, both cases show that evidence-conditioned gaps
remain strongly discriminative at the end of training, while bounded
modulation translates that contrast into controlled credit.

\subsection{Selective Credit Modulation Across Scales}

\begin{figure}[t]
\centering
\includegraphics[width=\linewidth]{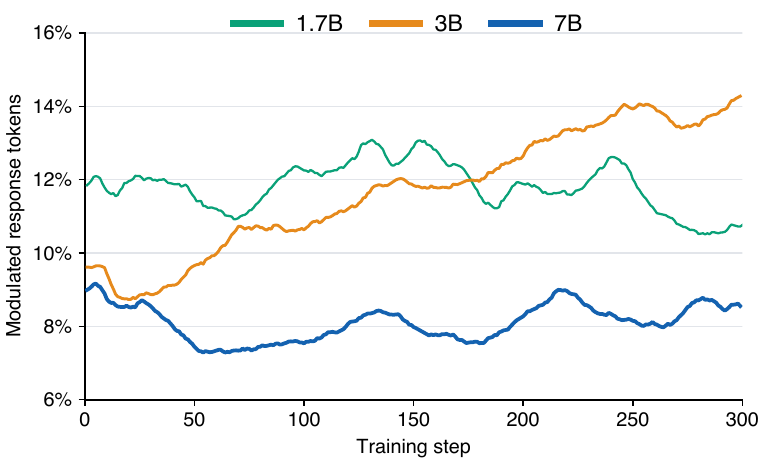}
\caption{Response-token coverage of EviSD's action-span mask across model scales.}
\label{fig:modulated_token_ratio}
\end{figure}

For each training step, modulation coverage is computed as
\[
\operatorname{Coverage}
=\frac{\sum_{i,k,t}m^{\mathrm{act}}_{i,k,t}}
        {\sum_{i,k,t}m^{\mathrm{resp}}_{i,k,t}}.
\]
EviSD concentrates privileged guidance on decision-critical search and answer
tokens rather than the full response. Across the complete training trajectories
of all three model scales, these actions comprise only 6.7\%--15.1\% of
response tokens (Figure~\ref{fig:modulated_token_ratio}). This consistently low
coverage demonstrates the token efficiency of selective credit modulation.

\subsection{Search Efficiency During Learning}

\begin{figure}[t]
\centering
\includegraphics[width=0.92\linewidth]{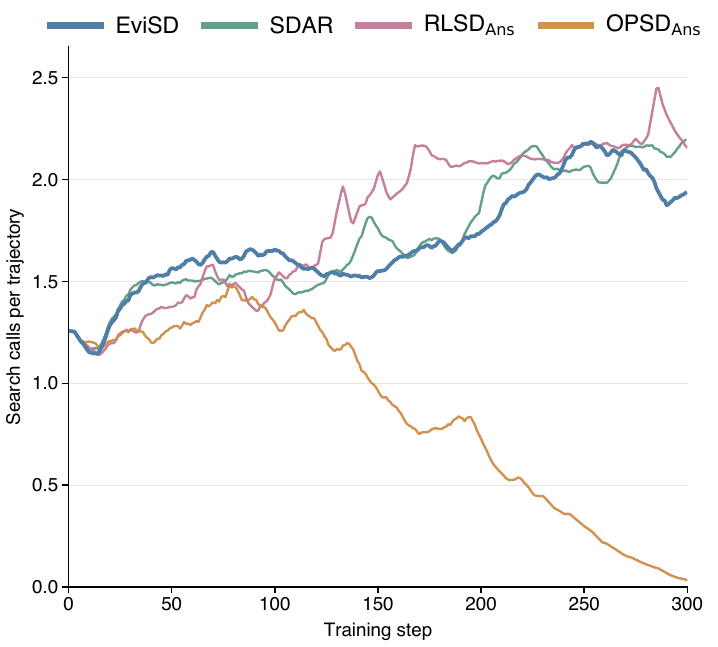}
\caption{Search calls per trajectory during Qwen2.5-7B training.}
\label{fig:search_calls}
\end{figure}

Figure~\ref{fig:search_calls} reports mean search calls per sampled trajectory.
The initial rise reflects acquisition of the interaction protocol. EviSD then
reduces its calls while maintaining the strongest on-policy success,
consistent with instance-specific evidence encouraging informative queries
and fewer redundant searches. OPSD's calls instead fall with its success,
marking search collapse rather than improved retrieval efficiency.

\subsection{Ablation Implementation Details}

Both context variants retain EviSD's action mask and bounded modulation.
\emph{Evidence + answer for all actions} concatenates both sources (using an
empty evidence field when unavailable), whereas \emph{Answer for all actions}
supplies only golden answers. Full-response modulation retains action routing
but replaces the action mask with the response mask, covering reasoning,
delimiters, and actions but not prompts or retrieved observations.

The response-wide auxiliary variant keeps the same routed teacher inputs but
replaces bounded action-token modulation with SDAR's sampled reverse-KL over
all valid response tokens. For states $s_t$ and $s_t^+$ on the sampled token,
\[
\begin{aligned}
\Delta_t&=\log\pi_T(y_t\mid s_t^+)-\log\pi_\theta(y_t\mid s_t),
\\
g_t&=\sigma(\beta\Delta_t),\\
\mathcal{L}_{\mathrm{SDAR}}&=\operatorname{Agg}_{t}[g_t\Delta_t].
\end{aligned}
\]
Teacher log probabilities and the gate are detached. With
$\lambda_{\mathrm{SDAR}}=0.01$ and $\beta=5.0$, we optimize
$\mathcal{L}_{\mathrm{GRPO}}+
\lambda_{\mathrm{SDAR}}\mathcal{L}_{\mathrm{SDAR}}$. 

\subsection{Hyperparameter Sensitivity}

\begin{figure}[H]
\centering
\includegraphics[width=0.90\linewidth]{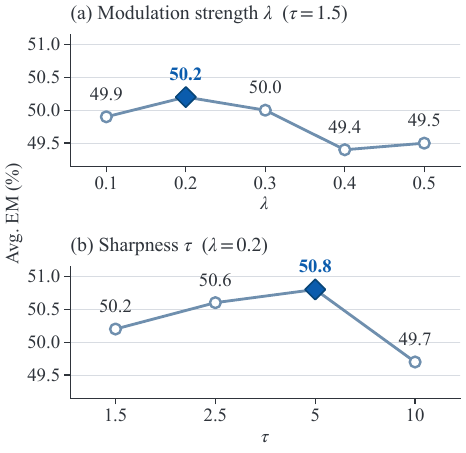}
\caption{Sensitivity to modulation strength $\lambda$ and sharpness $\tau$.}
\label{fig:hyperparameter_sensitivity}
\end{figure}

On Qwen2.5-7B-Instruct, we sweep $\lambda$ at $\tau=1.5$, then $\tau$ at
$\lambda=0.2$. Figure~\ref{fig:hyperparameter_sensitivity} shows stable average
EM for $\lambda\in[0.1,0.3]$ and degradation at larger values. Performance
improves with $\tau$ up to $5$ but drops at $10$, suggesting that excessive
sharpness over-concentrates the correction. We select $\lambda=0.2$ and
$\tau=5$ on 7B and transfer this configuration unchanged to all 3B and 1.7B
experiments.

\end{document}